\documentclass[conference]{IEEEtran}
\IEEEoverridecommandlockouts

\usepackage{algorithm}
\usepackage{algorithmic}
\usepackage{amsmath,amsfonts,amssymb,amsthm,mathtools}
\usepackage{caption}
\usepackage{cite}
\usepackage{graphicx}
\usepackage{tabularray}
\usepackage{textcomp}
\usepackage[dvipsnames]{xcolor}
\usepackage[caption=false]{subfig}

\def\BibTeX{{
    \rm B
    \kern-.05em{\sc i\kern-.025em b}
    \kern-.08em T
    \kern-.1667em\lower.7ex\hbox{E}
    \kern-.125em X}
}

\definecolor{mygray}{gray}{0.95}
\definecolor{mydarkgray}{gray}{0.85}

\begin{document}

\title{Learning Computational Efficient Bots\\with Costly Features}

\author{\IEEEauthorblockN{1\textsuperscript{st} Anthony Kobanda}
\IEEEauthorblockA{\textit{Ubisoft La Forge} \\
\textit{Ubisoft}\\
Bordeaux, France\\
anthony.kobanda@ubisoft.com}
\and
\IEEEauthorblockN{2\textsuperscript{nd} Valliappan C. A.}
\IEEEauthorblockA{\textit{Ubisoft La Forge} \\
\textit{Ubisoft}\\
Montréal, Canada\\
}
\and
\IEEEauthorblockN{3\textsuperscript{rd} Joshua Romoff}
\IEEEauthorblockA{\textit{Ubisoft La Forge} \\
\textit{Ubisoft}\\
Montréal, Canada\\
joshua.romoff@ubisoft.com}
\and
\IEEEauthorblockN{4\textsuperscript{th} Ludovic Denoyer}
\IEEEauthorblockA{\textit{Ubisoft La Forge} \\
\textit{Ubisoft}\\
Bordeaux, France\\
ludovic.denoyer@ubisoft.com}
}

\IEEEoverridecommandlockouts

\IEEEpubid{
    \makebox[\columnwidth]{979-8-3503-2277-4/23/$31.00~\copyright2023$ IEEE \hfill}
    \hspace{\columnsep}\makebox[\columnwidth]{ }
}

\maketitle

\IEEEpubidadjcol

\begin{abstract}

Deep reinforcement learning (DRL) techniques have become increasingly used in various fields for decision-making processes. However, a challenge that often arises is the trade-off between both the computational efficiency of the decision-making process and the ability of the learned agent to solve a particular task. This is particularly critical in real-time settings such as video games where the agent needs to take relevant decisions at a very high frequency, with a very limited inference time.

In this work, we propose a generic offline learning approach where the computation cost of the input features is taken into account. We derive the \textbf{Budgeted Decision Transformer} as an extension of the Decision Transformer that incorporates cost constraints to limit its cost at inference. As a result, the model can dynamically choose the best input features at each timestep. We demonstrate the effectiveness of our method on several tasks, including D4RL benchmarks and complex 3D environments similar to those found in video games, and show that it can achieve similar performance while using significantly fewer computational resources compared to classical approaches.\\

\end{abstract}


\section{Introduction}
\label{section:introduction}

Controlling real-time systems through intelligent methods is critical in many domains, notably including the deployment of bots\footnote{In a video game, a bot -- short for robot -- refers to any character operated by a computer program to simulate and, possibly, interact with a human player.} in video games.
Such agents often have to take real-time decisions in a dynamic and potentially stochastic environment by using input measures, or features, provided by sensors.
Usually features acquisition and decision making are separated, but for some applications they are deeply connected. Indeed they may share the same resources,
e.g., in video games whose GPU manages numerous processes at a very high frame rate, typically 60 Hz : rendering, game logics, AI systems, etc.

In real-time systems, the inference speed is critical and is prone to be influenced by the features the agent uses.
For instance, in autonomous car driving, it may be dependent on the time needed to gather and process images from cameras, or metrics from sensors, eventually planning how the different objects detected will move within the next few seconds, etc.
Similarly in video games, bots are usually nurtured by various information, some of them like their position, orientation, coming almost for free as they are naturally present in the game engine, and some others like raycasts\footnote{A raycast is the distance to the closest obstacle in a particular direction and is computed by iterating over rays to detect collisions in the 3D environment.} being much more expensive to compute. 
Using too many raycasts would not meet the real-time constraint, while not using any, or not the relevant ones, may exhibit unsatisfying behaviors.

\begin{figure}[t!]
    \centering
    \begin{minipage}[h!]{0.47\linewidth}
        \centering
        \includegraphics[width=1\linewidth]{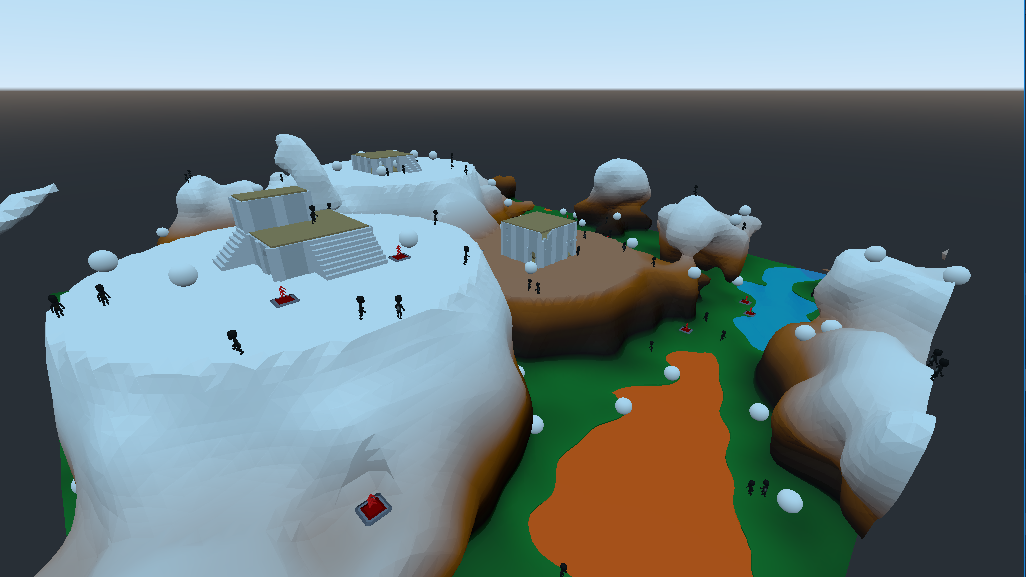}
        \subfloat{(A) View of a map.}
        \label{figure_godot_map}
    \end{minipage}
    \hspace{0.02\linewidth}
    \begin{minipage}[h!]{0.47\linewidth}
        \centering
        \includegraphics[width=1\linewidth]{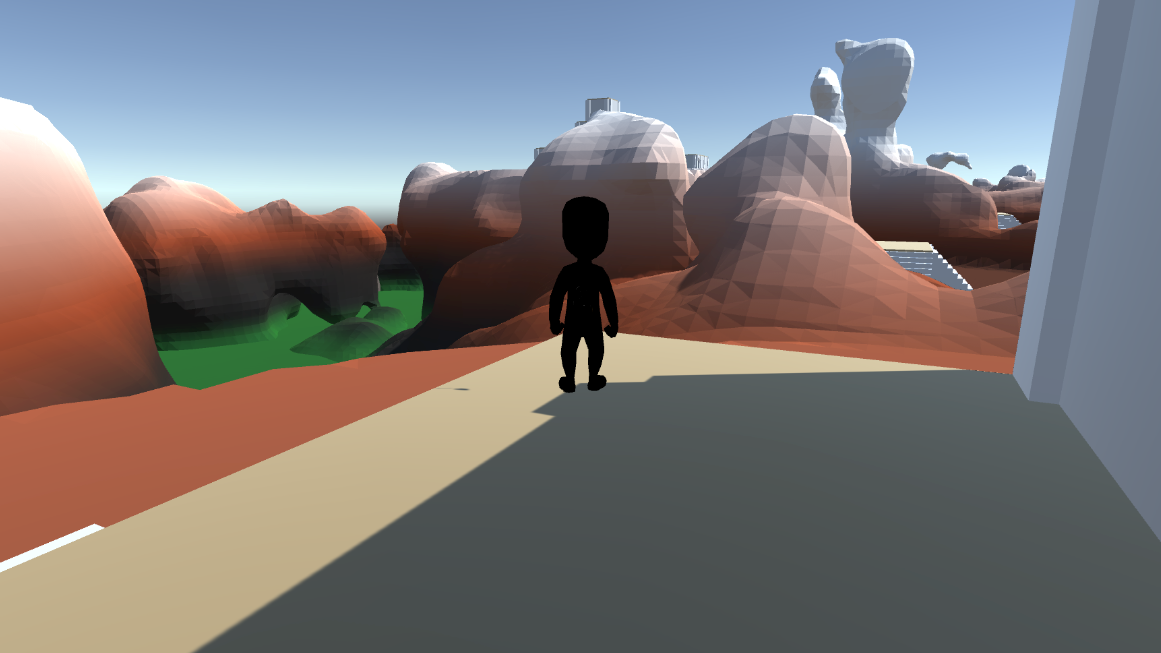}
        \subfloat{(B) Close-up of the agent.}
        \label{figure_godot_agent}
    \end{minipage}
\caption{Visualization of the Godot-based environments.
3D Navigation in video games requires the agent to move from a starting location to a given one in a complex map. The agent has access to basics information about its location, orientation, but also to raycasts that are expensive to compute.}
\label{figure_map}
\end{figure}

Deep Reinforcement Learning (DRL) is a natural
solution to learn agents in dynamic systems, generally achieved by training a neural network outputting actions from observations, with the objective to maximize a given cumulative reward over time.
Nevertheless, DRL algorithms usually do not constrain an agent on how much time it can spend to compute actions, and its speed is only a consequence of the choices made by the programmer (the features used, the neural network architecture implemented, and the hyper-parameters considered) and not of the algorithm itself.
If used to train bots in our video-game context, DRL however requires to be adapted to manage the trade-off between the inference speed and the agent quality.

Considering that computing features is the principal speed factor when executing bots in video games, we focus on \textbf{efficiently selecting features to perform well and fast}. 
This process of choosing an optimal subset of features is usually referred to as \textbf{(Cost-Sensitive) Active Features Acquisition} (AFA) in the machine learning literature (see the Section~\ref{section:related_work}) but rarely in the reinforcement learning setting, which usually assumes the observations as being freely provided by the environment.
When compared to the supervised learning one, where the goal is to find a good subset of features, it is a slightly different and more complicated problem in sequential decision-making: since the agent takes decisions sequentially, it becomes possible to choose different subsets of features at different timesteps e.g., a bot can learn to compute relevant raycasts only for critical frames, freeing time to compute other features when raycasts are not needed.

More particularly, we consider the \textbf{offline setting} where the agent is learned without interactions with the environment, directly from a dataset of collected episodes. Indeed, while AFA has been mainly studied in the online setting, we consider that, in a video game ecosystem, the offline setting is more realistic since the interaction with real AAA\footnote{In the video game industry, AAA is a classification for high-budget and eventually high-revenue games usually produced by well-known publishers.} games may be very slow, without the possibility to have multiple games running on a single machine. At the opposite, in this setting, it is usually possible to gather player/game interactions by tracking, resulting in large datasets of game traces that can be leveraged in order to learn efficient bots in an offline way. 

Our contributions are the following:
\begin{itemize}
\item We propose a formal definition of the cost-sensitive AFA problem with the objective to learn an agent both efficient in terms of collected reward and has a low computation cost by actively selecting features at each timestep.
\item We explain how classical offline learning losses can be extended to integrate the inference cost by specifying a maximum inference time a bot can use at each timestep.
\item We propose an implementation of these principles  using a Transformer architecture that allows the bot to focus on the best features at each timestep, resulting in a dynamic features acquisition model.
We detail the corresponding learning algorithm and other offline RL algorithm are instantiated with the same principles as comparisons.
\item We experiment on classical locomotion tasks from the D4RL benchmark, but also on realistic 3D Navigation tasks from a AAA prototype game where the computation speed of the learned agents is critical.
\end{itemize}

\section{Background and Problem Formulation}
\label{section:budgeted_learning}

\subsection{Offline Learning in Sequential Problems}
\label{subsection:offline_learning}

We consider a Partially-Observed Markov Decision Process
$\mathcal{M}=(\mathcal{S},\mathcal{O},\mathcal{A},\mathcal{T},\mathcal{R})$
composed of a state space $\mathcal{S}$,
an observation space $\mathcal{O}$,
an action space $\mathcal{A}$,
a transition function $\mathcal{T}:\mathcal{S}\times\mathcal{A}\rightarrow\mathcal{S}$
and a reward function $\mathcal{R}:\mathcal{S}\times\mathcal{A}\rightarrow\mathbb{R}$. An observation $o \in \mathcal{O}$ is defined as the set of all the features an agent, represented by a policy $\pi$, may use at each timestep to decide which action to take.
At any timestep $t$, a policy $\pi(\cdot|o_1,a_1,...,o_t)$ computes a distribution over $\mathcal{A}$ given the features of all the previous observations $o_1$ to $o_t$ and actions $a_1$ to $a_{t-1}$. When using memory-less architectures, the policy only uses the observation $o_t$ to choose the action to execute.

In addition, we consider a dataset $\mathcal{D}$ composed of episodes $(o_1,a_1,r_1),...,(o_T,a_T,r_T)$ acquired by one or many players or agents learned using all the features.
The episodes are of varying lengths, but for simplicity we consider a fixed one of $T$.
Given such a dataset, offline approaches usually define a loss function denoted $\Delta_\pi(\mathcal{D})$ which may capture the ability of the policy $\pi$ to predict similar actions to those in the dataset (imitation learning), conditioned to the reward-to-go (forecasting approaches) or a surrogate of the expected reward that the agent may obtain (offline RL approaches)\footnote{Different concrete algorithms are used as baselines in Section~\ref{section:experiments}.}.
In any case, the offline learning problem is to discover an optimal policy $\pi^*$ that minimizes the given loss function:
\begin{equation}
    \pi^* = \arg \min_\pi \Delta_\pi(\mathcal{D}).
\end{equation}

At inference time, the obtained policy can be evaluated by computing the expected cumulated reward $R(\tau)$ computed over the distribution of episodes $\tau$ generated by the policy $\pi^*$:
\[
 V_{\pi^*}=\mathbb{E}_{\tau \sim \pi^*}[R(\tau)],
\] 

Note that this setting assumes that the features used at train time are the same as the ones used at test time, and usually, the resulting policy uses all of them to predict the actions. 

\subsection{Active Features Acquisition in  Offline Learning}
\label{subsection:budgeted_offline_learning}

We now propose to define a different problem that we call \textit{Budgeted Offline Learning}: we consider that an observation $o_t$ is no longer the set of possible features the agent has to use, but \textbf{a subset of all the features it may decide to acquire}.
Indeed, while classical approaches ignore that the features composing the observation $o_t$ come with a computational cost, we aim at re-integrating this ignored step in the prediction process.

We decompose the policy into two sub-policies $\pi=(\hat{\pi},\tilde{\pi})$: the \textbf{acquisition policy} $\hat{\pi}$ samples a query $q_t$ that selects the features to acquire at time $t$. Then, depending on $q_t$, the agent will access the subset of acquired features in $o_t$ denoted $q_t(o_t)$,
and will decide which action to execute by computing the \textbf{action policy} $\tilde{\pi}(a_t|q_1(o_1),a_1,....,a_{t-1},q_t(o_t))$.
Note that the acquisition policy uses all the previously acquired observations as $\hat{\pi}(q_t|q_1(o_1),a_1,....,a_{t-1})$.
Hence, in our setting the agent will first sample a query $q_t$ and then an action $a_t$, based on previously acquired features, reintegrating the acquisition process before being able to act.

\paragraph{Feature Cost}
\label{paragraph:feature_costs} In addition to decomposing the policy, we consider the computation cost of the features.
A query $q_t$ is thus associated with a computation cost $c(q_t)$ which may be a cost in seconds, e.g., indicating how much time is needed to compute $q_t(o_t)$ at time $t$. Typically, this cost is higher when acquiring long-range raycasts in video games than short ones. 
We define the computation cost at time $t$ over trajectory $\tau$ as: 
\begin{equation}
\begin{aligned}
    C_t(\tau) = c(q_{t})
\end{aligned}
\label{equation:cost}
\end{equation}

This cost can thus be used as a feedback signal to encourage the model to limit its average computation cost at each step of the process as explained in the next paragraph. 

\paragraph{Limiting the Budget}
\label{paragraph:limiting_budget}
Given such a cost function, it is possible to define a constrained learning problem by integrating both the performance of an agent and its cost. We estimate the cost of a policy over the trajectories contained in our dataset $\mathcal{D}$ such that the resulting learning problem is:
\begin{equation}
\begin{aligned}
    \pi^* & = \arg \min_\pi \Delta_\pi(\mathcal{D}) \\
    \text{u. c.} &\ \forall \tau \in \mathcal{D}, \forall t, C_t(\tau)<\mathbf{C},
\end{aligned}
\label{equation:main_problem}
\end{equation}
where $\mathbf{C}>0$ is the maximal computation cost we allow the learned agent to have, and is manually set as a hyper-parameter (e.g no more than 1 ms).

This problem is a constrained optimization problem and various optimization techniques can be used to find a good minimum. 
We rely on the \textbf{Penalty Function} optimization \cite{10.5555/548530} which is an iterative algorithm that optimizes a sequence of unconstrained problems written as:
\begin{equation}
\phi_\pi(\tau,\mathbf{C})=\frac{1}{T} \cdot \sum\limits_{t=1}^{T}\max (C_t(\tau)-\mathbf{C},0),
    \label{equation:cost_loss}
\end{equation}
\begin{equation}
    \pi^* = \arg \min_\pi \sum\limits_{\tau \in \mathcal{D}} \Bigl( \Delta_\pi(\tau) + \gamma_k \cdot  \phi_\pi(\tau,\mathbf{C}) \Bigl),
    \label{equation:budgeted_loss}
\end{equation}
where $\gamma_k$ is the cost constraint weight at iteration $k$ of the algorithm. The underlying idea of Penalty Function is to define $\gamma_k$ such as $\gamma_k\geq0$ and $\gamma_{k+1} \ge \gamma_k$, with $\gamma_1=0$ in our case. By increasing the value of $\gamma_k$, the optimization algorithm enforces the constraints to be satisfied. (see Algorithm \ref{alg:example})

\section{Budgeted Decsision Transformers}

If the general approach presented in the previous section can be derived for different offline learning algorithms (see Section \ref{section:experiments}), we focus here on the extension of Decision Transformers (DT) to \textbf{Budgeted Decision Transformers} (BDT) to take into account costly features. Indeed, DT is a simple offline approach that can take into account multiple timesteps and that reaches high performance on standard offline RL benchmarks

\paragraph{Notations}
We consider that the observation space is composed of $m$ features denoted $o^1,...,o^m$. By features, we refer either to a scalar value or to a vector e.g., they can be a set of measures that are necessarily simultaneously acquired. Let us denote $f=(f_1,...,f_m)$ the acquisition costs of the different features that we assume as known at learning time.\footnote{Note that such cost values can be usually evaluated by running the environment multiple times and performing Monte-Carlo sampling.}

\begin{figure}[b!]
    \centering
    \includegraphics[height=4cm, width = 8.5cm,trim={0mm 0mm 0 0mm},clip]{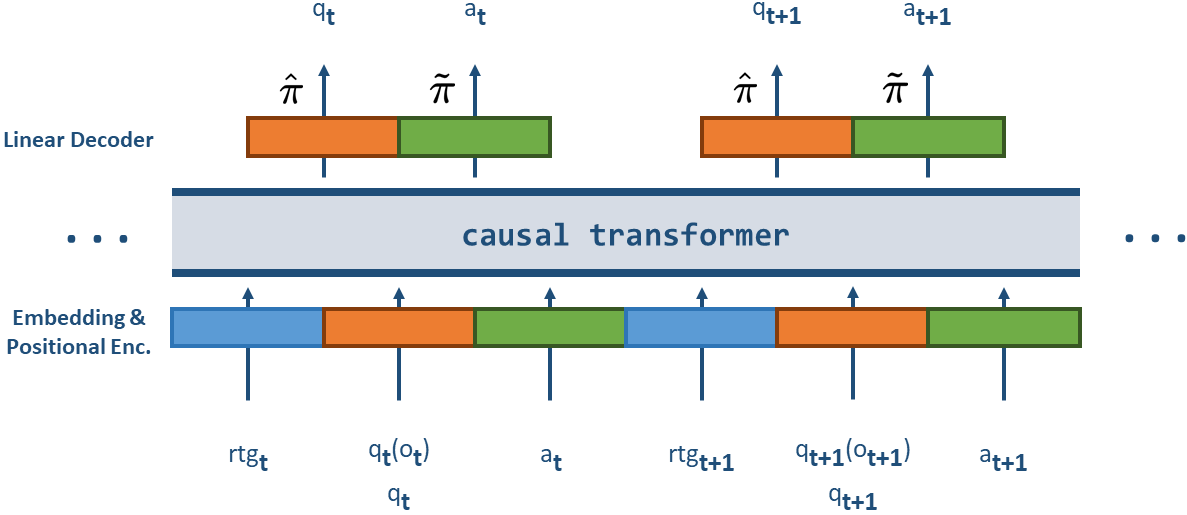}
    \caption{The Budgeted Decision Transformer Architecture: both the action and the acquisition policy are sharing the same transformer network.}
    \label{fig:bdt}
\end{figure}

\paragraph{Features Costs}
In such a context queries are regarded as binary masks $q_t \in \{0,1\}^m$ and the query results correspond to the set of the selected features: $q_t(o)=\{\ o^i\ |\ q^i_t=1\ \}$. The corresponding  \textbf{acquisition cost} is thus:
\begin{equation}
\begin{aligned}
c(q_t) = \frac{\langle\,q_t,f\rangle}{\left\lVert f \right\rVert_1}\in[0,1],
\end{aligned}
\end{equation}
which is a normalized cost to facilitate the tuning of the hyper-parameters of the algorithm.
This value provides insights about how computationally heavy a query is relative to the maximum query cost that may happen.

\textbf{Reward-to-go Conditioned Behavioral Cloning (RCBC):} BDT is based on the RCBC approach which consists in doing behavioral cloning using the reward-to-go denoted $rtg_t$ as an additional input to the policy. Known at train time, at inference time, the desired reward-to-go is specified by the user, and updated at each timestep depending on the immediate reward received by the agent \cite{bib_decision_transformers}. The corresponding loss function is:
\[
\Delta_\pi(\tau)=-\sum_{t=1}^T\log \pi(a_t|q_1(o_1),a_1,...,a_{t-1},q_t(o_t),rtg_t)
\]


\paragraph{Input Features} The selected features $q_t(o_t)$ at time $t$ are aggregated in a vector by concatenating the values $q_t^i \times o^i_t$ obtained by multiplying the mask value with the corresponding features. 
As an input, the action component receives vectors of real features values or zeros if not acquired.
To allow the model to disambiguate between zero-value features and not acquired ones, we also consider the query $q_t$ in the input:
\[
q_t(o_t)=\{q_t^1 \times o^1_t,...,q_t^m \times o^1_m,q_t^1,...,q_t^m\}
\]

In addition, we also use positional embeddings as in \cite{bib_transformers}.
\paragraph{Architecture}

The Budgeted Decision Transformer is an extension of the Decision Transformer with as inputs 
i) the past acquired features, 
ii) the past actions and 
iii) the (desired) reward-to-go. 
A difference between DT and BDT is the shared transformer architecture between two models: the acquisition policy $\hat{\pi}$ and the action policy $\tilde{\pi}$ as shown in Figure \ref{fig:bdt}. This model thus captures both distributions:
\[
\hat{\pi}(q_t|q_1(o_1),a_1,rtg_1,....,q_{t-1}(o_{t-1}),a_{t-1},rtg_{t-1})
\]
\[
\tilde{\pi}(a_t|q_1(o_1),a_1,rtg_1,....,q_{t-1}(o_{t-1}),a_{t-1},rtg_{t-1},q_t(o_t))
\]

Note that, since the action at time $t$ depends on the query at time $t$, this architecture loses its ability to compute the loss over all the timesteps simultaneously, resulting in a slower training speed. But this does not change the inference speed that has to be executed in the sequential order. Other faster architectures like transformer-XL \cite{dai2019transformerxl} could be used to speed up the model but it is not the focus of this paper.

\paragraph{Learning with Gradient Descent}
\label{paragraph:straight_through_estimator}

\begin{algorithm}[t!]
    \caption{Budgeted Learning Algorithm}
    \label{alg:example}
    \begin{algorithmic}
    
    \STATE {\bfseries Inputs:} Dataset $\mathcal{D}$, Cost constraint $\mathbf{C}$, Nb. of epochs $N$
    \STATE {\bfseries Initialize:} $\gamma_1 = 0$
    
    \FOR{k $\leftarrow$ 1 to $N$}
    \STATE $b \leftarrow sample(\mathcal{D})$ \COMMENT{{\small Sample a batch from the dataset}}
    \STATE $\tilde{\pi} \leftarrow \nabla_{\tilde{\pi}} \Delta_\pi(b)$ \COMMENT{{\small Action policy update using the classical loss}}
    \STATE $\hat{\pi} \leftarrow \nabla_{\hat{\pi}} \Delta_\pi(b) + \gamma_k \cdot \phi_\pi(b,\mathbf{C})$ \COMMENT{{\small Acquisition policy update using the constrained loss and the straight-through estimator}}
    \IF{$\phi_\pi(b,\mathbf{C})$}
    \STATE $\gamma_{k+1} \leftarrow \gamma_k + \epsilon$ \COMMENT{{\small If constraint is unsatisfied, increase $\gamma$}}   
    \ENDIF
    \ENDFOR   
    \end{algorithmic}
\end{algorithm}

An aspect we did not cover yet is that our acquisition policy $\hat{\pi}$ samples a binary query $q_t$ through a $m$-dimensional Bernoulli distribution. The training dataset contains the actions that have been taken allowing the learning of the action policy in a supervised way, but there is no information about the queries in the dataset and learning the acquisition policy cannot be solved using supervised learning. One option would be to rely on the online RL setting to discover $\hat{\pi}$ but would need interactions with the environment. We propose to learn to $\hat{\pi}$ together with $\tilde{\pi}$ by using the  \textbf{straight trough estimator} proposed in \cite{bib_straight_through_estimator}. This estimator makes the loss function of Equation \ref{equation:budgeted_loss} differentiable thus allowing gradient descent optimization.
The principle of this estimator is to produce binary outputs by sampling over a Bernoulli distribution but to back-propagate the gradient over the computed probabilities. It is implemented considering that:
\[
q_t = \mathcal{B}ernoulli(\tilde{q_t})+\tilde{q_t}-detach(\tilde{q_t})
\]
where $\tilde{q_t}$ is the vector of probabilities computed by $\hat{\pi}$. 

An other aspect is to encourage the policy to fulfill the desired constraint. The trade-off between the $\Delta$ loss and the cost constraint is controlled by a weight $\gamma$, but tuning the right value may be difficult. As explained in the previous section, we rely on Penalty Functions \cite{10.5555/548530} to solve this problem. Instead of increasing the value of $\gamma$ when the model has converged to a good solution, our implementation constantly increases $\gamma$ by a small step whenever the constraint is not satisfied in the current batch. In practice, such a strategy is stable and exhibits very good performance. It is illustrated in Algorithm \ref{alg:example}.

\section{Experiments}
\label{section:experiments}

\begin{figure}[t!]

    \centering
    \begin{minipage}[h!]{0.95\linewidth}
        \centering
        \includegraphics[width=1\linewidth]{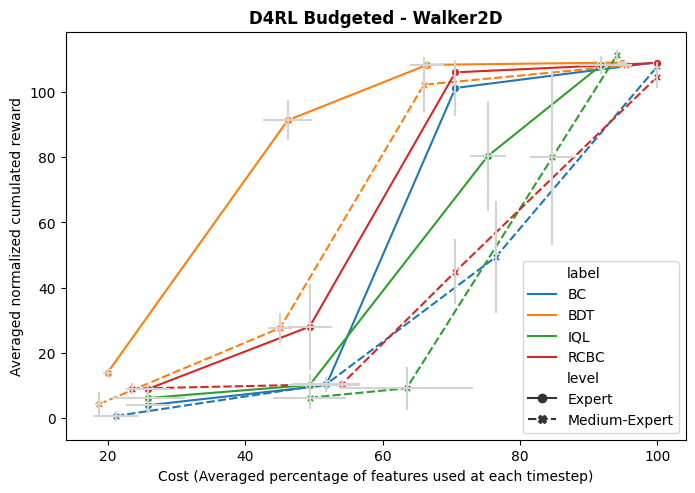}
        \subfloat{Performance/Cost on the Walker2D Environment}
        \label{HC}
    \end{minipage}
    \hspace{0.02\linewidth}
    \begin{minipage}[h!]{0.95\linewidth}
        \vspace{0.05\linewidth}
        \centering
        \includegraphics[width=1\linewidth]{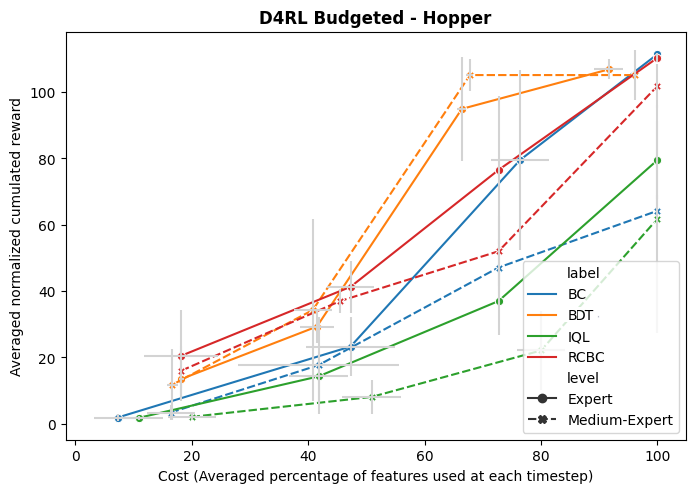}
        \subfloat{ Performance/Cost on the Hopper Environment}
        \label{W2D}
    \end{minipage}

\caption{Performance vs Cost of various models and various settings. Complete results are in supplementary material.}
\vspace{-0.3cm}
\label{fig:perf}
\end{figure}

\paragraph{Benchmarks} We now detail the different experiments made to validate our approach.
They rely on two families of environments: i) the \textbf{D4RL Benchmark} \cite{fu2020d4rl} suite that is composed of multiple locomotion environments each one associated with three different datasets of different qualities composed by episodes.
ii) The \textbf{3D Navigation} task, which is introduced in \cite{gamerlland}, is one of the real problems encountered in the video game industry where computation speed is critical\footnote{We provide a Godot re-implementation of this environment to only rely on open-source platforms in the supplementary material.}.
This setting contains two different 3D maps (easy and hard), both associated with two different datasets of medium and expert players generated as in D4RL, each dataset composed of 2,500 episodes. The agent can rotate, walk, jump, avoiding lava and water while trying to reach a goal (see supplementary material for more details). 

\paragraph{Baselines} The first models we evaluate are memory-less and based on simple multi-layers neural network (MLP) architectures that only consider observations at time $t$, as would a transformer of context size of 1. BC corresponds to behavioral cloning, RCBC is the reward-to-go version, and IQL corresponds to the implicit Q-learning \cite{kostrikov2021iql} algorithm that makes use of the immediate reward signal. All these three models can be extended with the additional cost term as in BDT\footnote{In IQL, there is no need to mask features at the level of the value and Q-functions, but only at the policy level which is the only component used at inference.} and are optimized with the same algorithm. BDT models are set with a context size of 20 as in the DT paper. 

\paragraph{Performance Metrics} For each set of experiment, we compute the average cost and average cumulated reward (or \textit{performance}) of the model and report these two values. All the results are averaged over three different training seeds, and performance has been computed over 256 episodes per training seed. We also report the standard deviation. Experiments have been made using eight computers equiped by 1080 GTX Nvidia GPUs and can be reproduced with a reasonably small cluster using the source code provided as supplementary material. This is an important aspect since it shows that the approach can be deployed in classical industrial contexts, and particularly at the level of single productions in video games.
We only report representative results in the main paper but all of them 
can be found in the supplementary material. 

\subsection{Performance/Acquisition cost trade-off}

In our first set of experiments, we focus on three D4RL locomotion tasks for which each feature is associated with a computation cost of $1$.
We train our models with cost constraints  $\mathbf{C}$ ($25\%$, $50\%$, $75\%$ and $100\%$) where $\mathbf{C}=25\%$ indicates that the model is constrained to select only one quarter of the possible features at each timestep. The results are presented in Figure \ref{fig:perf} -- see supplementary material for more complete results over all the environments and datasets, and a comparison with the literature. Note that there is always a difference between the desired cost and the one obtained, as the model is not always able to satisfy the constraint, and comparison between the have to be done at a similar real cost.

As a first result, we unsurprisingly see that the more a model is constrained, the lower the performance of the agent is. For instance, in a cost-free setting, all models are able to reach an average normalized cumulative reward of around 100 on the walker expert datasets while they barely reach half that performance when constrained to use only $50\%$ of the features. This can be observed in most of the cases and is unsurprising since all the observation features in the D4RL environments are important to solve each respective task. In that setting, limiting the number of used features naturally decreases the performance of the agent. One other remark is that handling the cost constraint is more difficult with IQL than with BC and RCBC since the IQL loss has a higher variance than the log likelihood losses in behavioral cloning. At last, the interest of conditioning on the reward to go decreases when using more and more expert traces since in that case, classical imitation learning models are sufficient enough. 

\begin{figure}[t!]

    \centering
    
    \begin{minipage}[h!]{1.0\linewidth}
        \centering
        \includegraphics[width=1\linewidth,height=2cm]{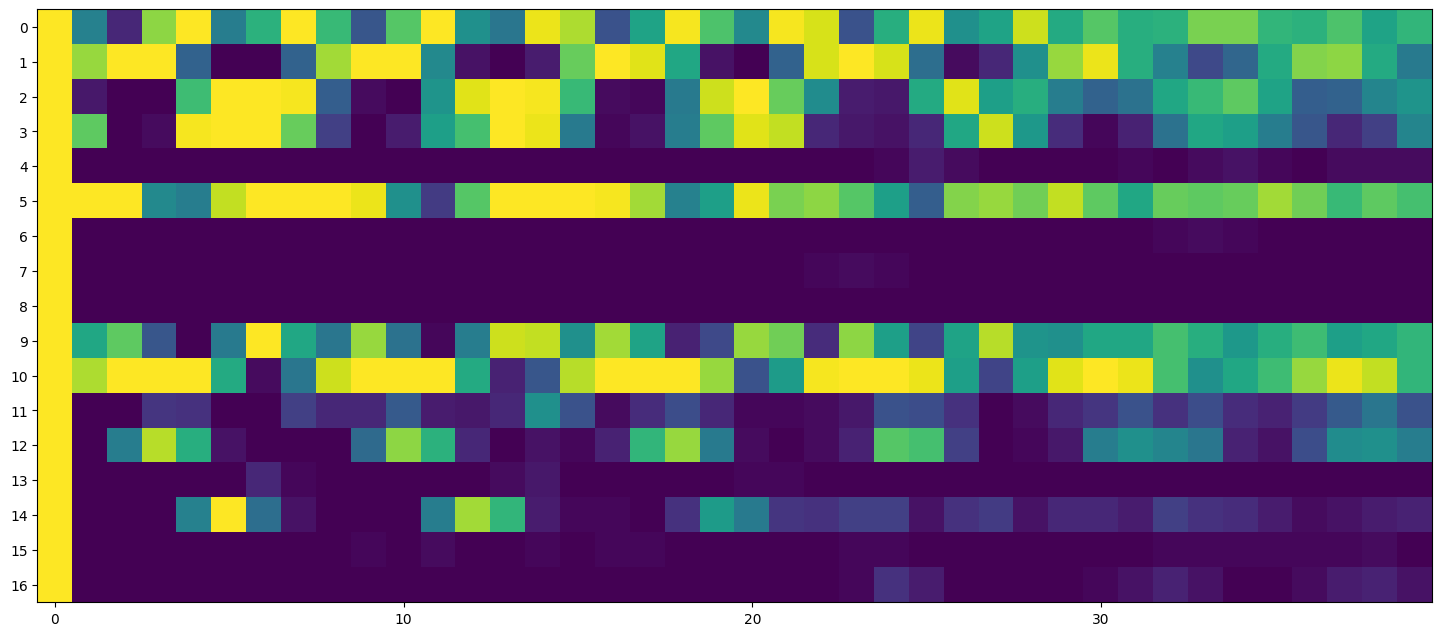}
        \subfloat{(A) Budgeted Decision Transformer}
    \end{minipage}
    
    \hspace{0.02\linewidth}
    
    \begin{minipage}[h!]{1.0\linewidth}
        \centering
        \includegraphics[width=1\linewidth,height=2cm]{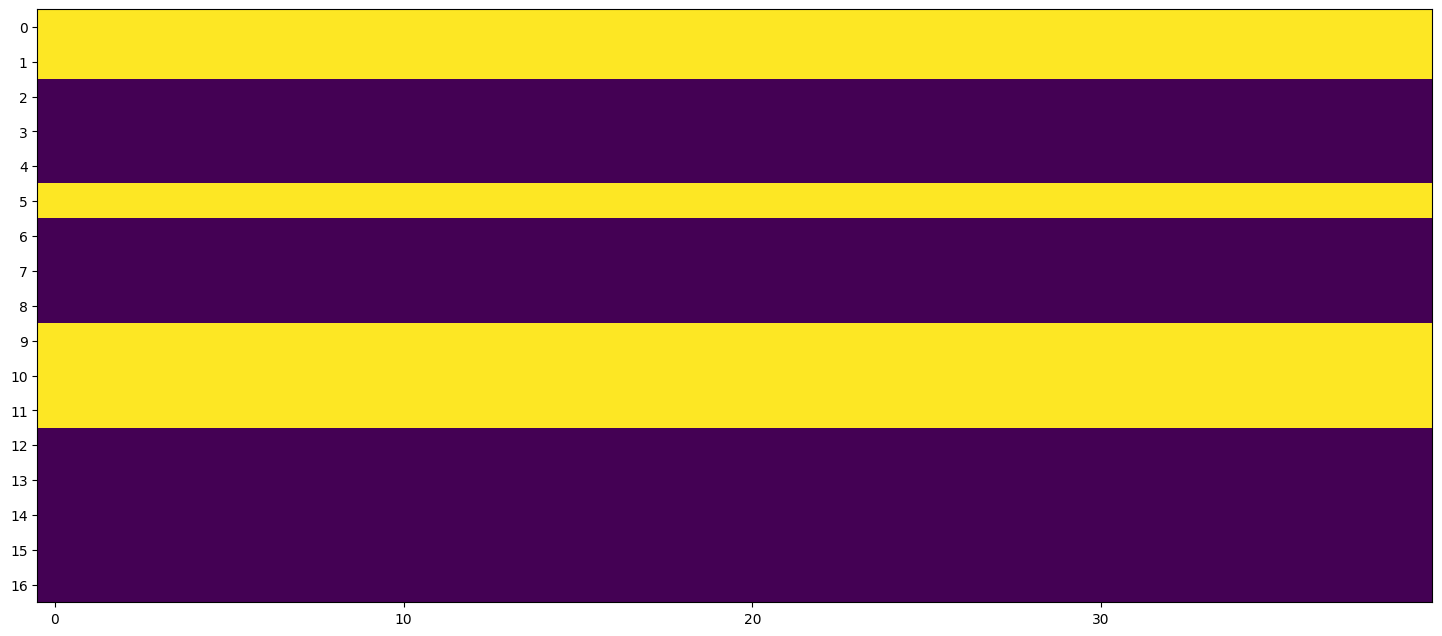}
        \subfloat{(B) RCBC (Memory-less setting)}
    \end{minipage}
    
    

    \caption{Features selected by various models on HalfCheetah-Expert when constraining to 25\% of the features. Each column corresponds to a timestep, and each row to one particular features. The brighter the color is, the highest the probability of selecting the features is (averaged on 256 episodes).}
    \label{fig:which_features_selected}

\end{figure}

Budgeted Decision Transformers are significantly better than the other models when using a similar constraint. It is particularly true when the constraint is high and the model has to use few features at each timestep. For instance, when using $60\%$ of the available features on Hopper medium-expert, BDT obtains a return of about 100, while other methods result in much lower performance.
This is confirmed in all the other environments and datasets. Indeed BDT can decide which features to acquire at each timestep instead of always acquiring the same ones. 
This dynamic selection process is illustrated in  Figure \ref{fig:which_features_selected}, comparing RCBC using a simple MLP and BDT.
This figure shows that BDT learns to cover more features than RCBC, and is able to constantly acquire particular relevant features, while switching to others at each timestep. BDT dispatches its budget over the timesteps as expected.  

We also compare the performance of BDT with one where the features are randomly sampled at each timestep instead of learning the acquisition policy (Table \ref{table:random}). The poor results obtained show how relevant are the features actively acquired.

\begin{table}[t]
    \scriptsize
    \centering
    \begin{tblr}{
        colspec = |c|c|c|c|c|c|c|c|,
        row{1} = {lime},
        row{2} = {mydarkgray},
        row{3-4} = {mygray},
        row{5} = {mydarkgray},
        row{8} = {mydarkgray},
        row{9-10} = {mygray},
    }
        \hline
        \textbf{Type} & \textbf{Environment} & \textbf{Cost Constraint} &  \textbf{BDT RANDOM}\\
        \hline
            
        & & baseline
            & 100.0 $\%$ $\vert$\ 92.7 $\pm$ 0.1 \\
        \hline
        Expert & HalfCheetah & 50 $\%$
            & {\color{brown} 48.8 $\pm$ 0.0 $\%$} $\vert$\ {\color{teal} 4.3 $\pm$ 1.1} \\
        & & 25 $\%$
            & {\color{brown} 24.0 $\pm$ 0.0 $\%$} $\vert$\ {\color{teal} 3.9 $\pm$ 0.9} \\

        \hline
            
        & & baseline
            & 100.0 $\%$ $\vert$\ 110.1 $\pm$ 1.6 \\
        \hline
        Expert & Hopper & 50 $\%$
            & {\color{brown} 48.6 $\pm$ 0.1 $\%$} $\vert$\ {\color{teal} 9.8 $\pm$ 10.8} \\
        & & 25 $\%$
            & {\color{brown} 23.7 $\pm$ 0.0 $\%$} $\vert$\ {\color{teal} 13.2 $\pm$ 3.3} \\

        \hline
            
        & & baseline
            & 100.0 $\%$ $\vert$\ 108.5 $\pm$ 0.8 \\
        \hline
        Expert & Walker2d & 50 $\%$
            & {\color{brown} 48.8 $\pm$ 0.0 $\%$} $\vert$\ {\color{teal} 15.8 $\pm$ 1.1} \\
        & & 25 $\%$
            & {\color{brown} 24.0 $\pm$ 0.0 $\%$} $\vert$\ {\color{teal} 5.8 $\pm$ 1.5} \\

        \hline
        
    \end{tblr}
    \caption{\textcolor{brown}{Average Cost} \& \textcolor{teal}{Average Performances} reached on D4RL datasets  ($N=1$) by BDT with features randomly sampled -- to compare with Figure \ref{fig:perf}.}
    \label{table:random}

\end{table}


\paragraph{Learning with costly (and noisy) features}

\begin{table*}[!ht]
    \scriptsize
    \centering
    \begin{tblr}{
        colspec = |c|c|c|c|,
        row{1} = {SkyBlue},
        row{2} = {mydarkgray},
        row{3-4} = {mygray},
        row{5-6} = {mydarkgray},
        row{7-8} = {mygray},
    }
        \hline
        \textbf{Cost Constraint} & \textbf{BDT using  only : No Noisy} & \textbf{BDT using only : No Noisy + Low Noisy} & \textbf{BDT using only : No Noisy + Low Noisy + High Noisy}\\
        \hline
        baseline
            & 100.0 $\%$ $\vert$\ 92.5
            & 100.0 $\%$ $\vert$\ 92.5
            & 100.0 $\%$ $\vert$\ 92.5 \\
        \hline
        221.0
            & \textcolor{brown} {190.7 $\pm$ 6.7 $\%$} $\vert$\ \textcolor{teal} {73.5 $\pm$ 14.8}
            & \textcolor{brown} {180.0 $\pm$ 10.2 $\%$} $\vert$\ \textcolor{teal} {89.0 $\pm$ 2.1}
            & \textcolor{brown} {190.0 $\pm$ 5.7 $\%$} $\vert$\ \textcolor{teal} {91.0 $\pm$ 0.1} \\
            & N = 56.0 $\%$ $\vert$\ L = 0.0 $\%$ $\vert$\ H = 0.0 $\%$
            & N = 31.0 $\%$ $\vert$\ L = 88.8 $\%$ $\vert$\ H = 0.0 $\%$
            & N = 30.5 $\%$ $\vert$\ L = 90.0 $\%$ $\vert$\ H = 22.5 $\%$ \\
        \hline
        110.5
            & \textcolor{brown} {82.9 $\pm$ 8.6 $\%$} $\vert$\ \textcolor{teal} {22.0 $\pm$ 2.8}
            & \textcolor{brown} {88.6 $\pm$ 4.1 $\%$} $\vert$\ \textcolor{teal} {88.5 $\pm$ 2.1}
            & \textcolor{brown} {97.9 $\pm$ 3.0 $\%$} $\vert$\ \textcolor{teal} {87.5 $\pm$ 0.7} \\
            & N = 24.5 $\%$ $\vert$\ L = 0.0 $\%$ $\vert$\ H = 0.0 $\%$
            & N = 5.0 $\%$ $\vert$\ L = 86.0 $\%$ $\vert$\ H = 0.0 $\%$
            & N = 5.5 $\%$ $\vert$\ L = 85.0 $\%$ $\vert$\ H = 42.5 $\%$ \\
        \hline
        22.1
            & \textcolor{brown}{ 8.8 $\pm$ 1.8 $\%$} $\vert$\ \textcolor{teal}{ 4.5 $\pm$ 0.7}
            & \textcolor{brown}{ 29.6 $\pm$ 6.4 $\%$} $\vert$\ \textcolor{teal}{ 24.5 $\pm$ 17.7}
            & \textcolor{brown}{ 24.6 $\pm$ 0.9 $\%$} $\vert$\ \textcolor{teal}{ 20.0 $\pm$ 2.8} \\
            & N = 2.5 $\%$ $\vert$\ L = 0.0 $\%$ $\vert$\ H = 0.0 $\%$
            & N = 0.0 $\%$ $\vert$\ L = 34.5 $\%$ $\vert$\ H = 0.0 $\%$
            & N = 0.0 $\%$ $\vert$\ L = 24.5 $\%$ $\vert$\ H = 22.5 $\%$ \\
            
        \hline
        
    \end{tblr}
    \caption{ \textcolor{brown}{Average Cost} \& \textcolor{teal}{Average Performances} reached by BDT on different noisy settings over the HalfCheetah environment using the Medium-Expert dataset. In this setup the features without any noise are worth 20, those with a low noise are worth 5 and the ones with a more important noise are worth 1. Since 17 features are originally available, the maximum cost constraint possible is $\mathbf{C}=17\times(20+5+1)=442$. We constraint our models to $50\%$, $25\%$ and $5\%$ of this value. We also provide the respective proportion of each features used (N, L, H) depending on their quality (not noisy, low noisy and high noisy).}
    \label{a1_budget_rewards_d4rl_noisy}

\end{table*}

To make the experiments more realistic and to measure the ability of our method to effectively identify relevant features when \textbf{features have different costs}, we introduce additional features in D4RL as noisy versions of the original ones.
We consider two additional sets of features: the \textbf{low-noise} features are copies of the D4RL features where a small Gaussian noise is added. 
The \textbf{high-noise} features use a stronger Gaussian noise making the resulting features more random. To model the fact that high quality information usually comes at a higher cost than low quality ones (e.g., a low resolution camera versus a high resolution one), we consider that the high-noise features have a cost of $1$, the low-noise features have a cost of $5$ and the original features have a cost of $20$. 

We consider three sets of environments where the agent has access to more or less noisy features. Table \ref{a1_budget_rewards_d4rl_noisy} 
illustrates the performance of BDT at different cost levels, the cost level being expressed as the value of $\mathbf{C}$ without normalization to allow a fair comparison between the different settings: it is the total cost of all the acquired featured. In addition, we provide the proportion of no-noise, low-noise and high-noise features used by the different models to better understand their behaviors and their respective features selection mechanism. 

First, at a given cost, we notice that the models learned over \textbf{no noisy + low noise} and \textbf{no noisy + low noise + high noise} settings outperform the models learned in the \textbf{no noisy} one, showing the ability of BDT to make use of noisy features that have a lower cost than the original ones. The same can be seen also when comparing the \textbf{no noisy + low noise + high noise} regularly outperforms the \textbf{no noisy + low noise} model, this model having access to cheaper features even if these features are more noisy. Since BDT is using previous timesteps, we interpret this phenomenon as an ability to rebuild meaningful information from the history of the noisy features. 

When looking at which features the models are using, it can be seen that, when constrained, they tend to select noisy and cheap features instead of the non-noisy and expensive ones.
Indeed, when possible, the models also continue to acquire few non-noisy features that are strongly relevant to solve the task -- e.g., at a cost constraint of $110.5$, $5\%$ of the non-noisy features are still acquired and used by the models. 

This set of experiments shows that our approach can benefit from features of poor quality if these features can be computed at a relatively low computation cost, opening the field to many possibilities when designing sensors for bots. 


\subsection{3D Navigation}

\begin{table}[!h]
\scriptsize
\centering
\begin{tabular}{|c|c|} 

    \hline
    Range of the raycast (in meters) & Compute Time (micro-seconds)  \\
    \hline
    1   & 2.58 \\
    5   & 4.27 \\ 
    10  & 5.14 \\ 
    25  & 5.37 \\ 
    50  & 5.49 \\ 
    100 & 7.9  \\
    \hline

\end{tabular}
\caption{The different real timing to compute a raycast on the 3D Navigation on a standard computer depending on the range of the raycasts, using the Godot engine.}
\label{fig:timegodot}
\vspace{-0.6cm}
\end{table}

The 3D Navigation task is a specific problem encountered in a multitude of video games environment, which is usually solved by using navigation meshes \cite{snook2000navmesh}.
Such a technique may produce unrealistic movements and displacement patterns, and so cannot be extended to complex map topologies \cite{deeprlnavigationaaa}.
In our setting, the agent is facing both free features (cost $= 0$) that are naturally computed by the video game (e.g. the agent position, its orientation, the goal position, etc...) but the bot can also compute additional features to capture finer information. 
Here, the agent has access to different raycasts launched in a particular direction, at a given range, associated with a given cost -- Table \ref{fig:timegodot}.
Each raycast returns both the distance to the closest obstacle in the range of the agent, but also the nature of the obstacle (ground, lava, water). A positive reward is given to the agent whenever it gets closer to the goal location (the goal location varying at each episode), while avoiding both lava and water (which instantly kill the agent) (see \cite{gamerlland}). We have evaluated both the RCBC and BDT architectures over two maps, and two datasets of trajectories per map of different qualities (medium and expert). 

\begin{figure}[t!]

    \centering
    \begin{minipage}[h!]{1.0\linewidth}
        \centering
        \includegraphics[width=1\linewidth]{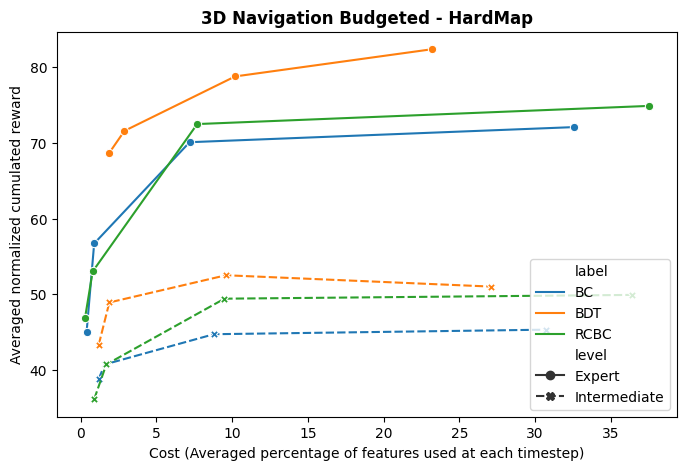}
        \label{HC}
    \end{minipage}

\caption{3D Navigation : Performance vs Cost of various models over the Hard Map using Godot engine.}
\vspace{-0.3cm}
\label{fig:perf2}
\end{figure}

The Table \ref{table:budget_godot_easymap} shows the cost/performance of the different methods.
First, by adding the cost constraint, we see that our models are able to reach comparable performance when compared with unconstrained models.
For example, on the easy map, the models constrained to use only $1\%$ of the raycasts achieve similar performances as the models that use all of the raycasts available.
This is due to the fact that using all the other features (agent location, orientation, etc.) are enough to find a path to the goal in this map.
On the hard map however, as expected, not using any raycasts provides bad performance (reward about $45$) in comparison to a model using all the features (reward is about $85$) -- see Figure \ref{fig:perf2}. But BDT is able to keep a reasonable performance spending only 20\% of the total cost on the expert dataset showing the ability of this method to detect which features are good trade-offs. When comparing BDT to RCBC, BDT tends to outperform RCBC at different cost levels, even if using a single MLP architecture in that case seems to be a reasonable solution, avoiding the complexity of transformer architecture.

\begin{table}[!t]
    \scriptsize
    \centering
    \begin{tblr}{
        colspec = |p{0.7cm}|c|c|c|c|,
        row{1} = {lime},
        row{2} = {mydarkgray},
        row{3-6} = {mygray},
        row{7} = {mydarkgray},
        row{12} = {mydarkgray},
        row{13-16} = {mygray},
        row{17} = {mydarkgray}
    }
        \hline
        \textbf{Env.} & \textbf{Const.} & \textbf{BDT} & \textbf{RCBC} \\
        \hline

        & base.
            & 100.0 $\%$ $\vert$\ R = 102.6 $\pm$ 0.6
            & 100.0 $\%$ $\vert$\ R = 101.9 $\pm$ 0.4 \\
        \hline
        & 50 $\%$
            & {\color{brown} 12.5 $\%$} $\vert $\ {\color{teal}R = 102.3 $\pm$ 0.3}
            & {\color{brown} 23.3 $\%$} $\vert $\ {\color{teal}R = 101.8 $\pm$ 0.9} \\
        Expert & 10 $\%$
            & {\color{brown} 9.7 $\%$} $\vert $\ {\color{teal}R = 102.3 $\pm$ 0.4}
            & {\color{brown} 8.0 $\%$} $\vert $\ {\color{teal}R = 102.3 $\pm$ 0.3} \\
        Easy & 1 $\%$
            & {\color{brown} 4.2 $\%$} $\vert $\ {\color{teal}R = 101.6 $\pm$ 0.5}
            & {\color{brown} 0.7 $\%$} $\vert $\ {\color{teal}R = 102.1 $\pm$ 1.1} \\
        & 0 $\%$
            & {\color{brown} 3.1 $\%$} $\vert $\ {\color{teal}R = 101.1 $\pm$ 0.0}
            & {\color{brown} 0.1 $\%$} $\vert $\ {\color{teal}R = 100.9 $\pm$ 1.7} \\
        
        \hline
            
        & base.
            & 100.0 $\%$ $\vert$\ R = 101.4 $\pm$ 0.7
            & 100.0 $\%$ $\vert$\ R = 101.9 $\pm$ 0.7 \\
        \hline
        & 50 $\%$
            & {\color{brown} 15.4 $\%$} $\vert $\ {\color{teal}R = 99.2 $\pm$ 0.5}
            & {\color{brown} 26.4 $\%$} $\vert $\ {\color{teal}R = 101.3 $\pm$ 0.2} \\
        Inter. & 10 $\%$
            & {\color{brown} 9.3 $\%$} $\vert $\ {\color{teal}R = 99.9 $\pm$ 0.1}
            & {\color{brown} 7.8 $\%$} $\vert $\ {\color{teal}R = 101.5 $\pm$ 0.2} \\
        Easy & 1 $\%$
            & {\color{brown} 2.6 $\%$} $\vert $\ {\color{teal}R = 99.4 $\pm$ 0.5}
            & {\color{brown} 1.0 $\%$} $\vert $\ {\color{teal}R = 99.0 $\pm$ 1.1} \\
        & 0 $\%$
            & {\color{brown} 2.1 $\%$} $\vert $\ {\color{teal}R = 98.7 $\pm$ 0.0}
            & {\color{brown} 0.4 $\%$} $\vert $\ {\color{teal}R = 99.0 $\pm$ 1.7} \\

        \hline

        & base.
            & 100.0 $\%$ $\vert$\ R = 82.7 $\pm$ 1.0
            & 100.0 $\%$ $\vert$\ R = 76.2 $\pm$ 1.0 \\
        \hline
        & 50 $\%$
            & {\color{brown} 27.1 $\%$} $\vert $\ {\color{teal}R = 82.4 $\pm$ 2.3}
            & {\color{brown} 36.4 $\%$} $\vert $\ {\color{teal}R = 74.9 $\pm$ 0.5} \\
        Expert & 10 $\%$
            & {\color{brown} 10.2 $\%$} $\vert $\ {\color{teal}R = 78.8 $\pm$ 2.4}
            & {\color{brown} 7.7 $\%$} $\vert $\ {\color{teal}R = 72.5 $\pm$ 5.1} \\
        Hard & 1 $\%$
            & {\color{brown} 2.9 $\%$} $\vert $\ {\color{teal}R = 71.6 $\pm$ 1.5} 
            & {\color{brown} 0.8 $\%$} $\vert $\ {\color{teal}R = 53.0 $\pm$ 5.5} \\
        & 0 $\%$
            & {\color{brown} 1.9 $\%$} $\vert $\ {\color{teal}R = 68.7 $\pm$ 4.7}
            & {\color{brown} 0.3 $\%$} $\vert $\ {\color{teal}R = 46.9 $\pm$ 2.2} \\

        \hline
        
        & base.
            & 100.0 $\%$ $\vert$\ R = 52.2 $\pm$ 1.7
            & 100.0 $\%$ $\vert$\ R = 50.2 $\pm$ 1.3 \\
        \hline
        & 50 $\%$
            & {\color{brown} 27.1 $\%$} $\vert $\ {\color{teal}R = 51.0 $\pm$ 0.5}
            & {\color{brown} 36.4 $\%$} $\vert $\ {\color{teal}R = 49.9 $\pm$ 1.3} \\
        Inter. & 10 $\%$
            & {\color{brown} 9.6 $\%$} $\vert $\ {\color{teal}R = 52.5 $\pm$ 0.9} 
            & {\color{brown} 9.5 $\%$} $\vert $\ {\color{teal}R = 49.4 $\pm$ 1.2} \\
        Hard & 1 $\%$
            & {\color{brown} 1.9 $\%$} $\vert $\ {\color{teal}R = 48.9 $\pm$ 0.4}
            & {\color{brown} 1.7 $\%$} $\vert $\ {\color{teal}R = 40.7 $\pm$ 0.5} \\
        & 0 $\%$
            & {\color{brown} 1.2 $\%$} $\vert $\ {\color{teal}R = 43.3 $\pm$ 3.0}
            & {\color{brown} 0.9 $\%$} $\vert $\ {\color{teal}R = 36.1 $\pm$ 0.3} \\

        \hline        
        
    \end{tblr}
    \caption{\textcolor{brown}{Average Cost} \& \textcolor{teal}{Average Performances} reached on the Godot environments. The table shows that using only very few raycasts is able to achieve high performance, and our method succeeds in identifying the relevant ones.}
    \label{table:budget_godot_easymap}

\end{table}

\section{Related Work}
\label{section:related_work}

{\setlength{\parindent}{0cm}

In supervised learning, feature selection has been the focus of many papers, using techniques such as $L_1$ regularization or sparse coding \cite{conf/icml/MairalBPS09}.
These methods have also been extended to the setting where features may have different cost values \cite{kachuee2019opportunistic,fainman2019online,janisch2019classification}.
In supervised learning, an alternative approach is to model the selection process as a reinforcement learning problem.
For example, \cite{lizotte2012budgeted} formulated the budgeted learning problem as a Markov Decision Process, \cite{rasoul2021feature} uses TD learning, or even in \cite{contardo2016sequential} which considers different features cost values.\linebreak
In these works, the reinforcement learning aspects are focused on the sequential nature of the features selection process, modeled through a policy choosing what to acquire step by step. 
But we emphasize that these methods essentially target supervised learning tasks, and consequently only target prediction over single data points.\\  

A more general approach is to consider not only the selection process, but the complete discovery of the architecture of the prediction model -- Neural Architecture Search (NAS). 
Different methods have been proposed \cite{alshubaily2021efficient, nagy2021improving, mazyavkina2021optimizing} where RL is used as a way to learn the architecture of the neural network, but these methods are still time-intensive. 
Some recent research work addresses this drawback by performing simultaneous architecture search while learning the network parameters as shown in \cite{bib_learning_budgeted_network} and the adaptive model proposed by \cite{bib_stochastic_nas}. 
But again, the objective is to do NAS for supervised learning over classical datasets, without studying sequential decision problems. 
In the context of learning agents through RL, NAS has been recently studied \cite{zoph2016neural, nagy2021improving} by extending classical NAS methods using online-RL methods instead of supervised learning ones. 
But, the objective is to learn an efficient architecture, and the NAS is not constrained by costs of any nature. If we focus on the features selection problem which is the main topic of our paper, only few papers have been proposed in the RL community.
For instance \cite{pmlr-v28-nguyen13} studies sparse coding for model-based RL. The closest work to our is certainly  \cite{sanner:ecai08} where the authors extended Q-learning by adding a $L_1$ regularizer to select the right features. The limit of this paper is that they study only one online algorithm (Q-learning) with very simple neural network architectures while our contribution is proposed for offline RL, by using transformer architectures.\\

The Transformer architecture \cite{bib_transformers} has been recently invested in the reinforcement learning domain  \cite{heess2015memory,kumar2020adaptive,NEURIPS2022_b2cac94f, parisotto2020stabilizing} showing great performance. Particularly, the decision transformer \cite{bib_decision_transformers} and the trajectory transformer \cite{janner2021offline} leverage the use of expert data and learn in an offline way. Our work is directly inspired from decision transformers where we add a budget constraint to limit the inference cost.\\

Finally, in RL, the active features acquisition problem has been tackled through different ways.
For instance, \cite{rl1} proposes a setting where obtaining the reward information is costly in an online setting, the agent being able to ask for the reward or not at each timestep. 
\cite{rl4} considers that one can ask for zero or all the features but does not consider individual features.
The closest works to ours are \cite {rl2} and \cite{rl3} that attack a similar problem in an online setting.
For instance \cite{rl2} uses a VAE to rebuild missing features resulting in a complex method, while \cite{rl3} is not using long-range architectures as we do.





\newpage
\section{Conclusion and Discussion}
\label{sec:conclusion}

We have proposed Budgeted Decision Transformers as an efficient way to learn agents that not only decide which actions to take but also which features to acquire to limit the computation cost, resulting in fast agents. The efficiency of this approach has been demonstrated in various environments and settings from a classical offline RL benchmark (D4RL) to a concrete industrial problem, i.e., 3D navigation in a AAA video game prototype.
We want to highlight first that we do not consider the computation cost of the transformer architecture and only compute the acquisition cost over features since the computation time of the neural network itself greatly varies depending on the setting, multiple agents eventually being batched together for a faster inference. Moreover, in our concrete case, the computation time of the network is lower than the one of the features.\\
 
Second, an important aspect of the paper is the offline setting. Indeed, doing budgeted learning in an online setting would necessitate to learn many bots for multiple cost constraints, each learning being made by interacting with the environment. It would be at a very high computation price that is unrealistic in many companies where frugality is required. Focusing on the offline learning setting allows us to proceed differently: a single agent can be learned through online learning using all possible features. Once a good agent has been found, it is possible to generate a big dataset of episodes and to learn the budgeted models over this dataset at different cost constraints, as it is done in distillation methods. By doing so, budgeted learning is made possible on small infrastructures and encourages industrialization of RL models in video games.

\bibliography{sample}
\bibliographystyle{ieeetr}

\end{document}